\documentclass[10pt,twocolumn,letterpaper]{article}

\usepackage[pagenumbers]{cvpr} 

\usepackage{booktabs} 
\usepackage{graphicx}
\usepackage{amsmath}
\usepackage{amssymb}
\usepackage{booktabs}
\usepackage[title]{appendix}
\usepackage[utf8]{inputenc}
\usepackage{graphicx}
\usepackage{threeparttable}
\usepackage{tabularx}
\usepackage{xcolor}
\usepackage{multirow}
\usepackage{soul}
\usepackage{amsmath,amssymb,amsfonts,dsfont,bm,mathrsfs}
\usepackage{caption}
\usepackage{xcolor}
\usepackage{booktabs,multirow,adjustbox}
\usepackage{float}
\newcommand\blfootnote[1]{%
  \begingroup
  \renewcommand\thefootnote{}\footnote{#1}%
  \addtocounter{footnote}{-1}%
  \endgroup
}

\usepackage[pagebackref,breaklinks,colorlinks]{hyperref}

\begin{document}

\title{FEAT: Face Editing with Attention}
\author{
  Xianxu Hou\textsuperscript{1},
  Linlin Shen\textsuperscript{1}\thanks{Corresponding author},
  Or Patashnik\textsuperscript{2},
  Daniel Cohen-Or\textsuperscript{2},
  Hui Huang\textsuperscript{1} \\
  \textsuperscript{1}Shenzhen University \quad
  \textsuperscript{2}Tel Aviv University \\
}

\twocolumn[{
\renewcommand\twocolumn[1][]{#1}
\maketitle
\begin{center}
  \includegraphics[width=1\linewidth]{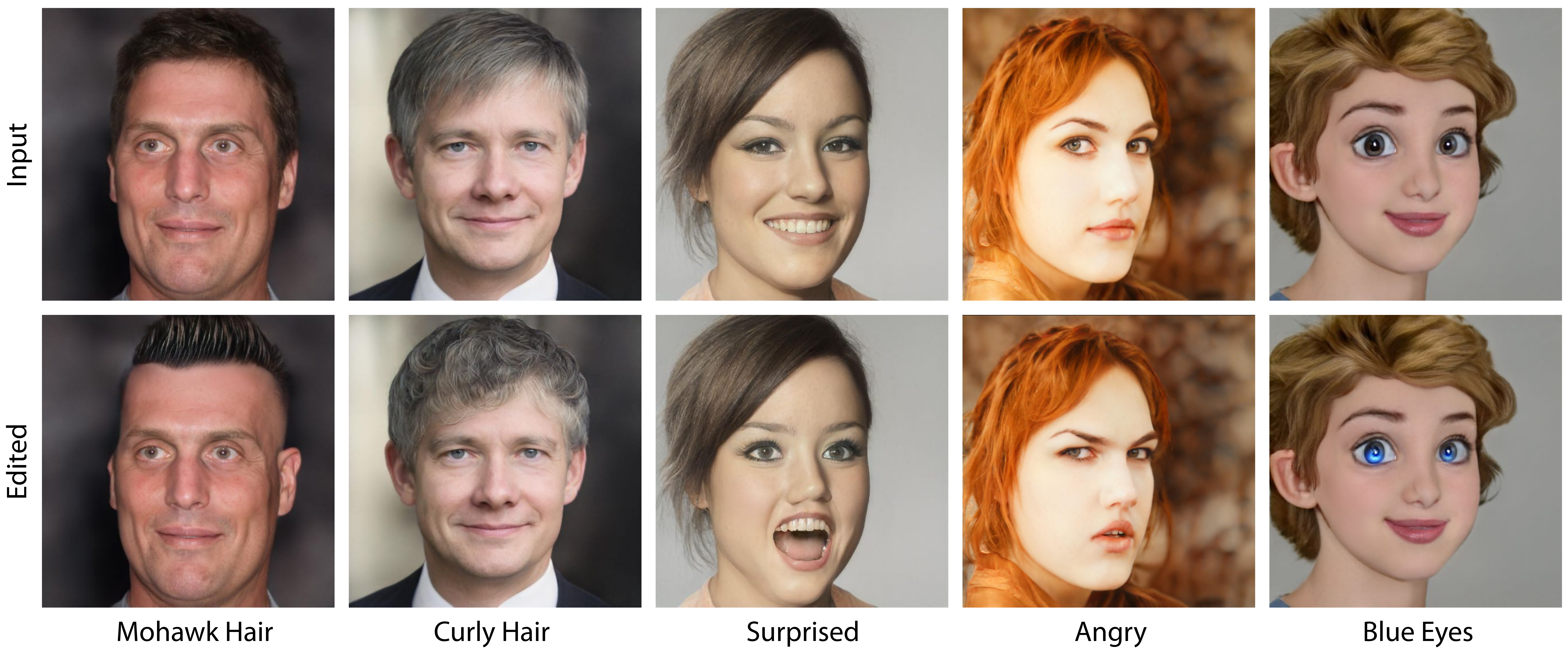}
  \vspace{-7pt}
  \captionsetup{type=figure,font=small}
  \caption{Manipulating local facial attributes based on text descriptions. The input and edited images are shown in the first and second row, respectively. The corresponding text prompts are shown at the bottom.}
  \label{fig:teaser}
  \vspace{5pt}
\end{center}
}]

\blfootnote{
\hspace{-12pt}
$^*$Corresponding author}

\begin{abstract}
Employing the latent space of pretrained generators has recently been shown to be an effective means for GAN-based face manipulation. 
The success of this approach heavily relies on the innate disentanglement of the latent space axes of the  generator. However, face manipulation often intends to affect local regions only, while common generators do not tend to have the necessary spatial disentanglement. 
In this paper, we build on the StyleGAN generator, and present a method that explicitly encourages face manipulation to focus on the intended regions by incorporating learned attention maps. During the generation of the edited image, the attention map serves as a mask that guides a blending between the original features and the modified ones. The guidance for the latent space edits is achieved by employing CLIP, which has recently been shown to be effective for text-driven edits. We perform extensive experiments and show that our method can perform disentangled and controllable face manipulations based on text descriptions by attending to the relevant regions only. Both qualitative and quantitative experimental results demonstrate the superiority of our method for facial region editing over alternative methods.
\end{abstract}

\section{Introduction}
\label{sec:intro}

Recent years have witnessed the significant progress of Generative Adversarial Networks (GANs)~\cite{goodfellow2014generative}. 
Specifically, StyleGAN~\cite{karras2019style,karras2020analyzing,karras2020training,aliasfreeKarras2021}, one of the most celebrated GAN frameworks, can produce high fidelity human face images with unmatched photo-realism. Furthermore, it has been shown that StyleGAN provides semantically rich latent space, which allows us to edit images in a semantically meaningful manner.


Numerous GAN-editing works use direction in the latent space to edit an image~\cite{shen2020interpreting, harkonen2020ganspace, shen2020closed, wu2021stylespace}. However, using such a direction for manipulation heavily relies on the assumption that the latent space is perfectly disentangled. Furthermore, these works typically require manual tuning of hyperparameters such as manipulation strength and disentanglement magnitude.
Other works take an alternative approach and train a network that predicts a per-image offset in the latent space for an intended edit~\cite{hou2020guidedstyle, abdal2020styleflow, alaluf2021matter, patashnik2021styleclip}. These methods neither assume perfect disentanglement in the latent space, nor require manual tuning of the edit magnitude.

The most evident challenge in the aforementioned works is how to achieve a disentangled edit, that is, editing the intended attribute without affecting other attributes. In many cases, the disentanglement is achieved naturally, without explicit enforcement. However, some cases are more challenging and require special means. Previous works achieve disentanglement by constructing a spatially-disentangled representation of the latent space~\cite{kafri2021stylefusion}, or by operating in other latent spaces, for example, StyleSpace~\cite{collins2020editing, wu2021stylespace,xu2021generative}. However, these methods heavily rely on the innate disentanglement of the latent space axes, which do not always have a spatial disentanglement.

\begin{figure*}[!t]
\begin{center}
   \includegraphics[width=1\linewidth]{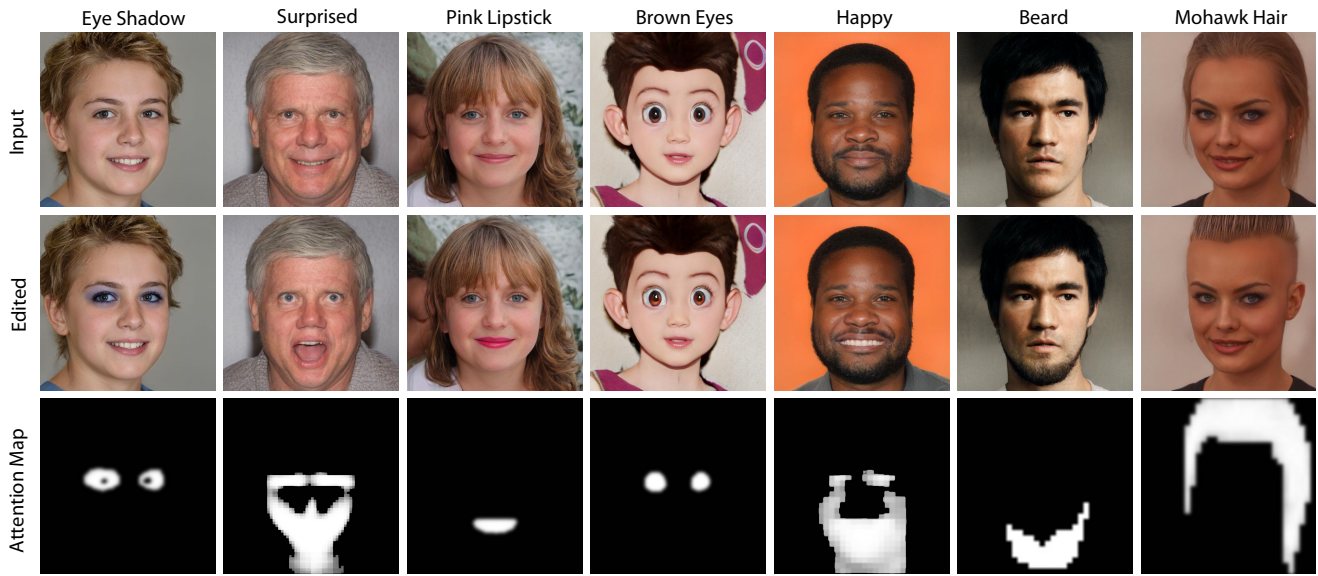}
\end{center}
   \caption{Edited results of our methods using different text prompts. The input and edited results are shown in the first and second row, respectively. The corresponding attention masks are shown in the third row.
   }
\label{fig:attention-map}
\end{figure*}

In this paper, we present a method that explicitly encourages the edit to be focused on the intended region by incorporating the \emph{learned} attention map.
Specifically, given a latent code and a target edit, we train a network that  predicts an offset in the latent space. During training, we learn an attention map by accounting for the features of all layers. The attention module that computes the masks is also guided by the edited attribute. The attention map is applied on the features of a target layer during the generation of the edited image. Blending the masked features with the original ones leads to the manipulation of the intended regions only. To guide the edit, we use CLIP~\cite{radford2021learning}, which has been shown to be effective for a wide range of image manipulations, 
and thus enables a simple and intuitive user interface (see Fig.~\ref{fig:teaser} and Fig.~\ref{fig:attention-map}).

To validate the effectiveness of the attention maps, we perform extensive experiments and show that we are able to obtain image changes in the intended regions by only using text descriptions. Moreover, we analyze the choice of the layers in which the attention maps are applied and show that it is necessary to apply them in the feature space. Finally, we compare our method with previous methods and show the advantage of using such attention maps.

\section{Related Work}
\label{sec:related}


\paragraph{GAN Latent Space Manipulation.}
Generative Adversarial Networks have demonstrated great potential in learning various high-level semantics from observed data in the latent space, which provides an intuitive way to control the image generation process with different semantic specifications. An active line of research on image editing is to directly manipulate the latent code of a pretrained GAN model. Early works~\cite{hou2017deep,radford2015unsupervised} observed interpretable vector arithmetic, which enables easy editing of different visual concepts in the learned latent space. Most notably, the discovery of such interpretable directions have received much research attention with the advancement of StyleGAN. Recent works have sought to identify a semantically meaningful path in a supervised manner, which requires a large number of annotated images~\cite{abdal2020styleflow} or attribute predictors on the predefined semantics~\cite{shen2020interpreting,hou2020guidedstyle}. 

Two recent works~\cite{harkonen2020ganspace,shen2020closed} have considered unsupervised identification of interpretable controls over image synthesis by using a closed-form factorization method or applying PCA in the latent space. While these approaches do not require supervision, they can only find limited semantic directions, which are selected manually and with extensive effort. Recently, domain adaptation methods ~\cite{song2021agilegan,jang2021stylecarigan,zhu2021mind} that build upon StyleGAN have also been proposed to achieve style transfer. Several works~\cite{collins2020editing,kafri2021stylefusion,kim2021exploiting} have focused on local semantically-aware edits to a target output image by encoding the local semantics of images into the StyleGAN latent space. Additionally, in order to support real image manipulations, GAN inversion~\cite{abdal2019image2stylegan,zhu2021improved,zhu2020indomain,richardson2020encoding,tov2021designing,alaluf2021restyle,roich2021pivotal,alaluf2021hyperstyle,xia2021gan} has been adopted to inversely project a given image into the latent space of a pretrained GAN generator, and then the obtained codes can be edited in a semantically meaningful manner.

\paragraph{Text-based Image Synthesis and Manipulation.}
GANs have also been used to synthesize natural images based on text descriptions. For example, in some works, a conditional GAN has been used to generate bird and flower images by explicitly taking a language embedding as input. StackGAN~\cite{zhang2017stackgan} and StackGAN++~\cite{zhang2018stackgan++} use multiple generators and discriminators to further improve the quality of generated images. AttnGAN~\cite{xu2018attngan} uses an attentional generative network to select the condition at the word level for synthesizing fine-grained details in different subregions of the image. The semantic consistency and image fidelity can be further improved through carefully designed network architecture and training objectives~\cite{zhu2019dm,qiao2019mirrorgan,yin2019semantics,li2019object,cheng2020rifegan}. More recently, Gal et al.~\cite{gal2021stylegannada} present a text-based zero-shot domain-adaptation technique. 

Beyond text-to-image generation, manipulating face images using text descriptions has recently become popular with the advancement of StyleGAN. TediGAN~\cite{xia2021tedigan} achieves text-based image generation and manipulation by mapping the image and text descriptions into a common StyleGAN latent space. StyleCLIP~\cite{patashnik2021styleclip} uses Contrastive Language-Image Pretraining (CLIP) model~\cite{radford2021learning} to develop a text-based interface for StyleGAN image manipulation. While these approaches can provide a certain level of text-based global attribute control, the local linguistic information of key words is not really attended. As a result, unwanted changes or artifacts may be produced. Our goal is to manipulate the semantics of a given image by automatically attending to the relevant regions.

\paragraph{Attention Learning.}
Attention learning has been successfully introduced in many applications in natural language processing and computer vision, such as image captioning~\cite{xu2015show}, object localization~\cite{oquab2015object}, visual question answering~\cite{yang2016stacked} and machine translation~\cite{bahdanau2015neural}. Learning attention can also encourage more realistic image generation and manipulation. For example, SAGAN~\cite{zhang2019self} considers the non-local relationships in the feature space by incorporating a self-attention module into the GAN framework. Moreover, several unsupervised image-to-image translation techniques~\cite{chen2018attention,yang2019show,alami2018unsupervised,mokady2019masked} use attention network to predict spatial attention maps of images, demonstrating significant improvements in the quality of the translated images. A recently proposed method~\cite{bau2021paint} enables users to manipulate a semantics of a GAN generated image at a given location, however it requires to manually specify the locations. In addition, a proposed facial structure editing method~\cite{wu2021coarse} aims to remove the double chin in portrait images relying on neck masks, which are extracted by a pretrained facial parsing model. By contrast, in our work we achieve controllable face manipulation by automatically attending to the regions of interest using only full-text descriptions.

\section{Method}
\label{sec:method}

\begin{figure*}[t!]
\centering
\includegraphics[width=\linewidth]{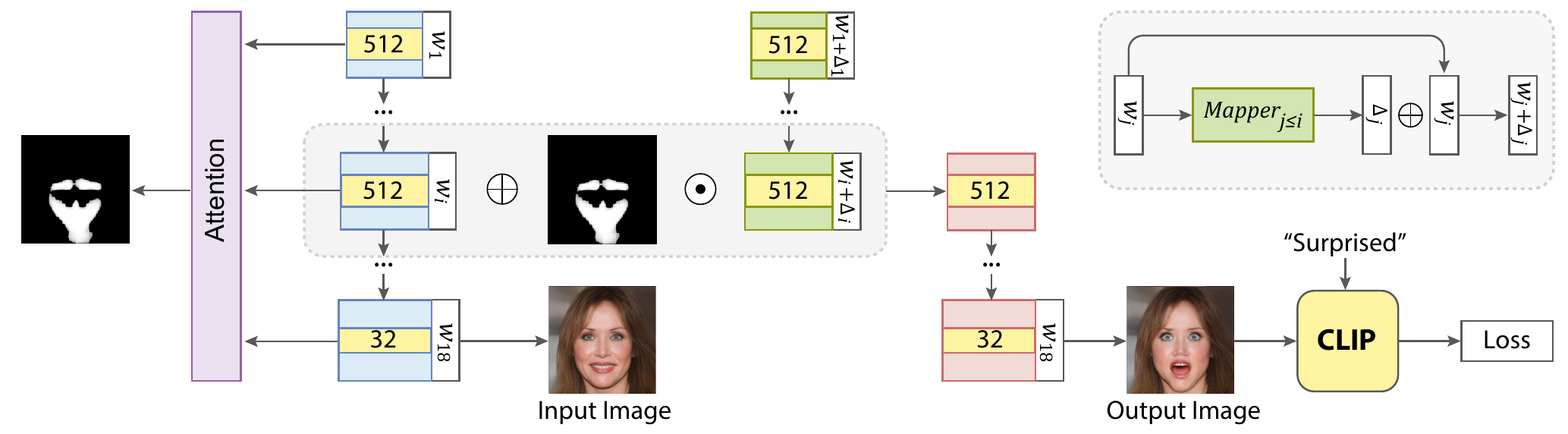}
\caption{Overview : given an image generated by $w=(w_1, \cdots, w_{18}) \in \mathcal{W^+}$ (the features are in light blue, where the number of channels are marked in the yellow band) and guided by a text prompt, we train mappers for style codes $w_j$ for all $j \leq i$ that predict corresponding offsets $\Delta_j$. We refer to the image generated by the modified $w_j + \Delta_j$ as the mapped image. In addition, we train an attention module (in light purple) that combines the features of all layers into a single attention map (on the left). Then, the $i^{th}$ layer features of the original image and the mapped one are blended using the learned attention map.}
\label{fig:overview}
\end{figure*}

\subsection{Preliminaries}
StyleGAN2 is the current state of the art model for high-resolution image generation owing to its unique generator architecture. Instead of directly feeding the input latent code $z \in \mathcal{Z}$ to the network, it uses a mapping network $f$ to transform $z$ to the intermediate latent code $w = f(z)$, which is then transformed to style codes that control the layers of synthesis network $g$ by modulating each layer's convolutions. In layer $i + 1$ of $g$, the modulated convolution is applied on the features of layer $i$, denoted by $S_i(w_i)$.  The output image is obtained by transforming StyleGAN2's features to RGB features via toRGB layers. StyleGAN2's intermediate latent space, $\mathcal{W}$, has been shown to be semantically meaningful and disentangled. These properties make StyleGAN2 very effective for image manipulation. It has become a common practice to edit images by traversing in StyleGAN2's latent space.

\subsection{Overview}
Our method, FEAT, employs StyleGAN2's latent space to edit an image (see Fig.~\ref{fig:overview}). FEAT is built on previous methods that train a mapper $h$ that learns an offset in the latent space to edit the image. To attain a disentangled edit, and to edit the intended regions only, FEAT learns an attention map to blend the features obtained by the source latent code with the features of the shifted latent code.
To guide the edit, we employ CLIP~\cite{radford2021learning}, which allows for using text to learn the offset and generate the attention map. 

\begin{figure}[t!]
\centering
\includegraphics[width=\linewidth]{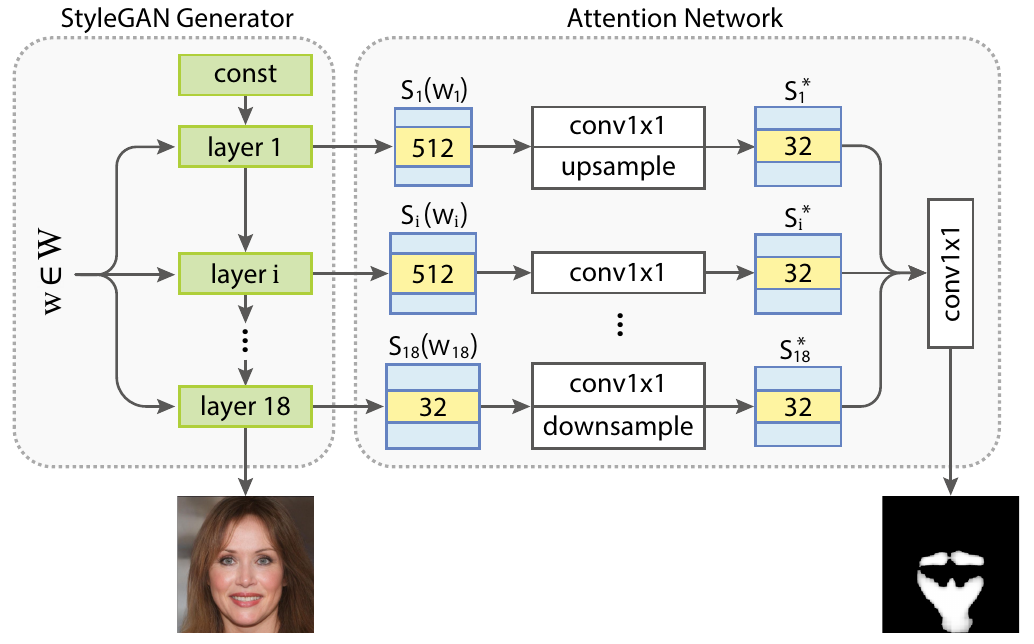}
\caption{The architecture of our attention network $f_{att}$. The left part is the fixed StyleGAN2 generator for feature extraction. The right part is the attention network used to produce the attention masks.}
\label{fig:attention-network}
\end{figure}

\subsection{The Mapper and Attention Module}
Our mapper network $h$ is implemented as an MLP that receives a latent code $w$ and outputs an offset. The edited code is obtained by predicting the residual from the original latent code:
\begin{equation}
    w_{edited} = w + \alpha \cdot h(w),
\end{equation}
where $\alpha$ is a hyper-parameter used to balance between the original and edited data. 

To edit the intended regions only, we employ an attention network $f_{att}$ that learns to produce an attention map $m$, in which each pixel has a probability value between 0-1. 
The attention map is applied on the $i^{th}$ layer features, $S_i(w_{edited})$, obtained during the generation of the image corresponding to $w_{edited}$. The attention map has a single channel, and the spatial dimensions of $S_i$.
We apply $m$ to the features via an element-wise product in each channel to create the attended area: $m \odot S_i(w_{edited})$ (green features in Fig.~\ref{fig:overview}). Similarly, the unattended area is created by $(1-m) \odot S_i(w)$ (blue features). Thus, the final features (red features) are defined as follows:
\begin{equation}
\label{eq:attention}
\Tilde{S_i} = m \odot S_i(w_{edited}) + (1-m) \odot S_i(w).
\end{equation}
Having obtained these blended features, the final edited image is attained by applying layers $> i$ on $\Tilde{S_i}$, with $w$ modulating the corresponding convolutions.
During testing, we threshold the learned attention maps by only containing pixels exceeding a user-defined threshold $\tau$, which we set to 0.8 in our experiments.

Inspired by the recent work~\cite{zhang2021datasetgan}, we exploit the features of all StyleGAN2 layers for the training of the attention network $f_{att}$. Fig.~\ref{fig:attention-network} illustrates the attention network architecture. We first use multiple $1 \times 1$ convolutional layers to reduce the number of channels of the feature maps $S_1(w_1), ..., S_i(w_i), ..., S_{18}(w_{18})$. For simplicity, all the feature maps are reduced to the same number of channels (32 by default in our implementation). The resulting feature maps are resized to the resolution of $S_i(w_i)$ and concatenated to a single deep feature map $S^* = [S_1^*, ..., S_i^*, ..., S_{18}^*]$. Here, $S_i^*$ denotes the transformed feature map in the $i^{th}$ layer, and $S^*$ is then fed to another $1 \times 1$ convolution to produce a single channel feature map as the attention mask. A sigmoid layer is used at the end to ensure the output mask is bounded in the $(0, 1)$ interval.

\subsection{Training Objectives}

\paragraph{Semantic Consistency Loss.}
First, we adopt a text-image matching network to define a semantic consistency loss, which guides the image manipulation to match the provided text descriptions. In particular, given the text description $t$ and the pretrained CLIP, our semantic consistency loss is defined as:
\begin{equation}
\begin{aligned} 
\label{eq:clip_loss}
\mathcal{L}_{clip} &= D_{clip}(I^*, t)
\end{aligned}
\end{equation}
where $D_{clip}$ denotes the cosine distance between the image and text embeddings extracted with CLIP, and $I^*$ is the blended output image. Note that the features of the blended image outside the mask remain intact. Thus, the CLIP model is encouraged to focus the optimization on the attended area only, and the attention module encourages semantic consistency.

\paragraph{Attention Map Regularization.}
To encourage the attention network to focus on a more compact region related to the text descriptions rather than the whole image, we further introduce an attention map regularization as: 
\begin{equation}
\label{eq:att_loss}
\mathcal{L}_{att} = \frac{1}{HW} \sum_{i=1}^{H} \sum_{j=1}^{W} m_{ij},
\end{equation}
where $H$ and $W$ are the width and height of the learned attention map, respectively, and $m_{ij}$ is the pixel value at the location $(i, j)$.

\paragraph{Total Variation Loss.}
To encourage spatial smoothness in the produced attention map $m$, we adopt a total variation penalty as:
\begin{equation}
\label{eq:tv_loss}
\mathcal{L}_{tv} = \sum_{i, j} \|m_{i+1, j} - m_{i, j}\|_2 + \| m_{i, j+1} - m_{i, j}\|_2.
\end{equation}

\paragraph{Latent Loss.}
To explicitly preserve visual attributes of the input images, we minimize the $L_2$ distance between the original latent code $w$ and the transformed code $w_{edited}$ as:
\begin{equation}
\label{eq:l2_loss}
\mathcal{L}_{l_2} = \|w - w_{edited}\|_2.
\end{equation}

\paragraph{Overall Training Objective.}
Our full objective is the weighted sum of these four losses:
\begin{equation}
\label{eq:all}
\mathcal{L} = \mathcal{L}_{clip} + \lambda_{att} \mathcal{L}_{att} + \lambda_{tv} \mathcal{L}_{tv} + \lambda_{l_2} \mathcal{L}_{l_2}, 
\end{equation}
where $\lambda_{att}$, $\lambda_{tv}$ and $\lambda_{l_2}$ are the hyper-parameters to control the relative importance of each component.

\section{Results and Evaluation}
\label{sec:results}

\subsection{Experimental Settings}
\label{sec:setting}
\paragraph{Datasets.}

We use the StyleGAN2 model pretrained on FFHQ dataset~\cite{karras2019style} as our generator. The FFHQ dataset is a $1024 \times 1024$ resolution high-quality image dataset of human faces. We adopt e4e~\cite{tov2021designing} to project the test set of CelebA-HQ~\cite{karras2018progressive} for real image manipulation. We also use StyleGAN2 model fine-tuned on cartoon faces for evaluation (see, for example, the rightmost column in Fig.~\ref{fig:teaser}). In addition, we perform the editing on real faces collected from the Internet.

\begin{figure*}[!tb]
\begin{center}
   \includegraphics[width=\linewidth]{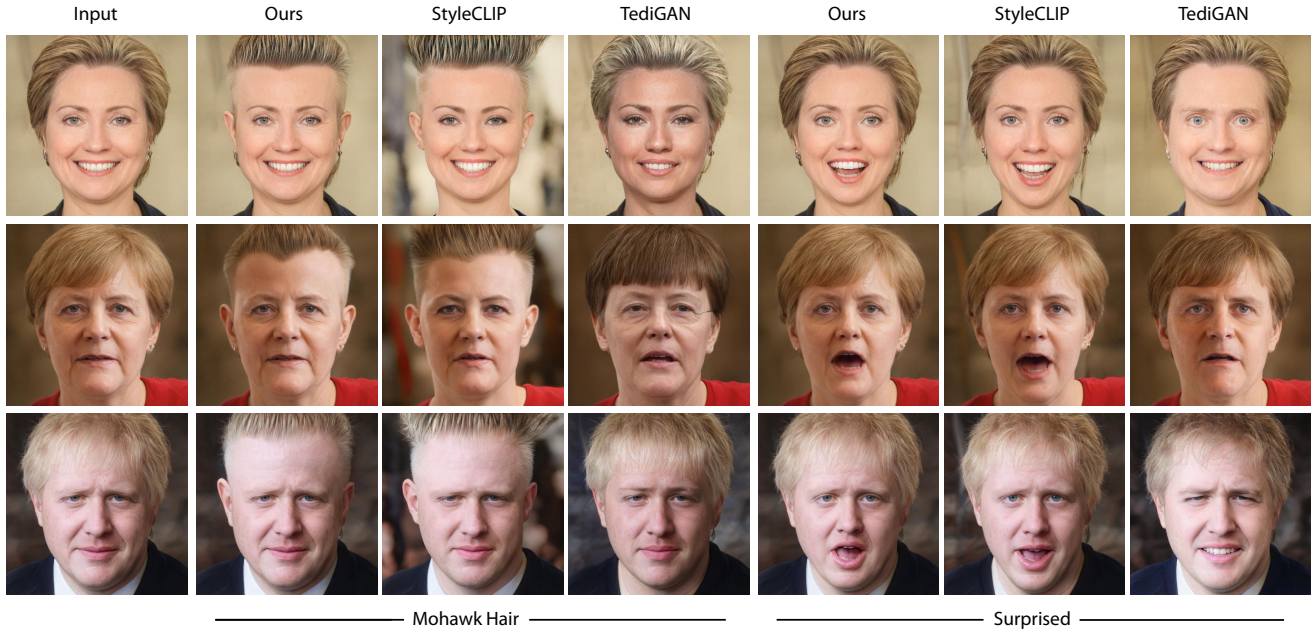}
\end{center}
   \caption{Visual comparison of our method with TediGAN and StyleCLIP. The driving descriptions are indicated at the bottom.}
   \label{fig:compare1}
\end{figure*}

\begin{figure}[!tb]
\begin{center}
   \includegraphics[width=\linewidth]{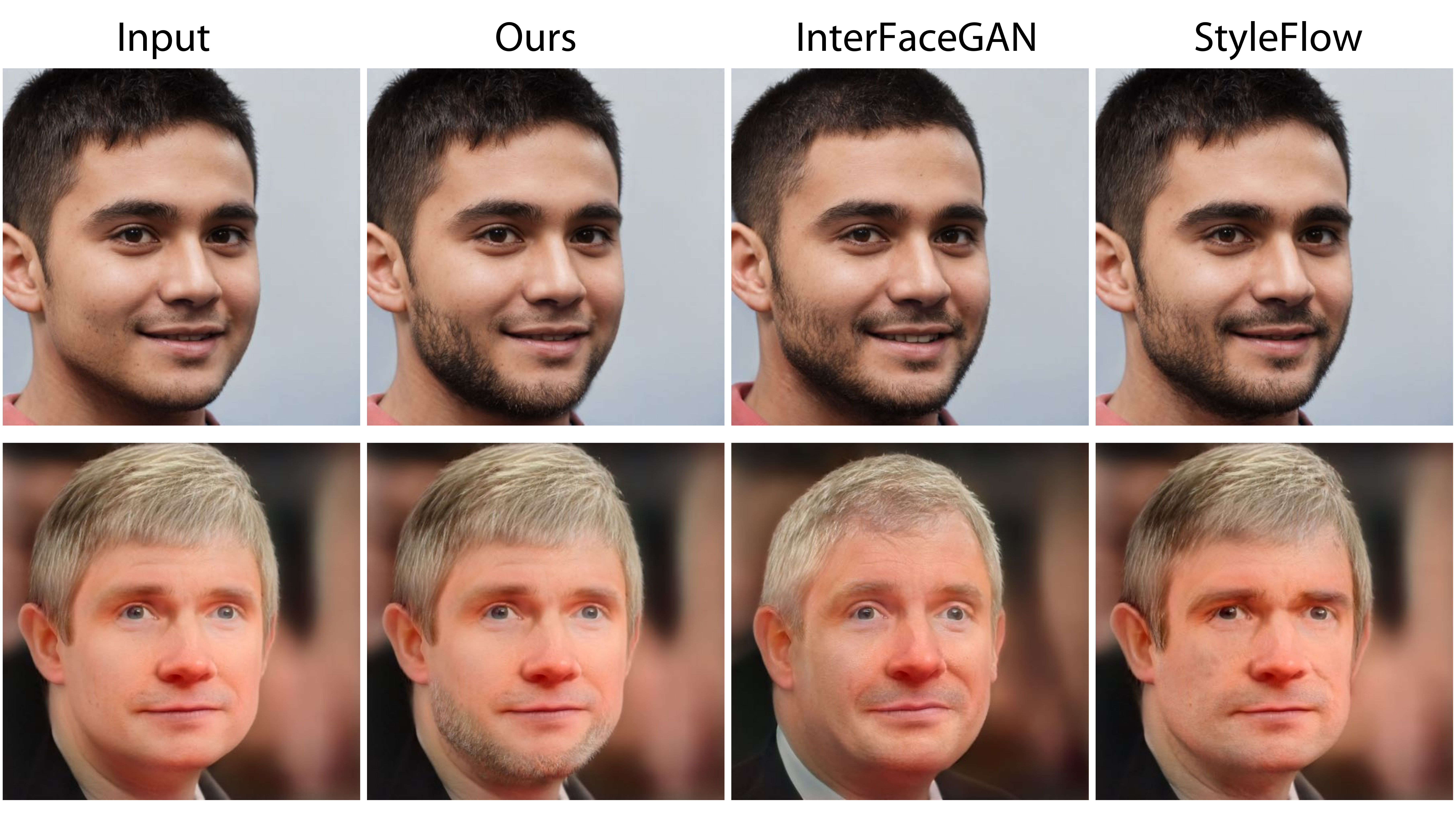}
\end{center}
   \caption{Visual comparison of our method on "beard" manipulation with InterFaceGAN and StyleFlow.}
\label{fig:compare2}
\end{figure}

\paragraph{Baselines.}
TediGAN~\cite{xia2021tedigan} and StyleCLIP~\cite{patashnik2021styleclip} are closely related works, as both can achieve semantic face manipulation based on text descriptions. In addition, we use InterfaceGAN~\cite{shen2020interpreting} and StyleFlow~\cite{abdal2020styleflow} as our baselines, which control face manipulations along predefined semantic directions.

\paragraph{Evaluation Metrics.} We evaluate both the visual quality and identity preservation using quantitative metrics. Fr\' echet Inception Distance (FID)~\cite{heusel2017gans} is used to measure the discrepancy between the edited and original faces. We adopt the face recognition model FaceNet~\cite{schroff2015facenet} to extract the embeddings of tested images and use Cosine Similarity (CS) and Euclidean Distance (ED) to quantify the identity preservation.

\paragraph{Training Details.}
We adopt a three-layer MLP with 512 hidden units and 512 output units to build our mapper network. Image sampling is performed by randomly drawing from a normal distribution from the $\mathcal{Z}$ space, which is then mapped into the intermediate latent space $\mathcal{W}$. The output images are resized to $224 \times 224$ before feeding them to CLIP (using the pretrained ViT-B/32 weights). We use Adam optimizer to train the mapper and attention network at a learning rate of 0.0001. We use a batch size of 2 on one Tesla V100 32GB GPU. The maximum number of iterations is set to 20,000. Note that the pretrained StyleGAN2 is fixed. We set $\lambda_{att} = 0.005$, $\lambda_{tv} = 0.00001$, $\lambda_{l_2} = 0.8$, and $\alpha = 0.1$. We edit the first eight layers for facial structure manipulation such as hairstyle and expression, and edit all the 18 layers when performing local color manipulation, such as that for the hair and eyes.

\begin{figure}[!tb]
\begin{center}
   \includegraphics[width=0.98\linewidth]{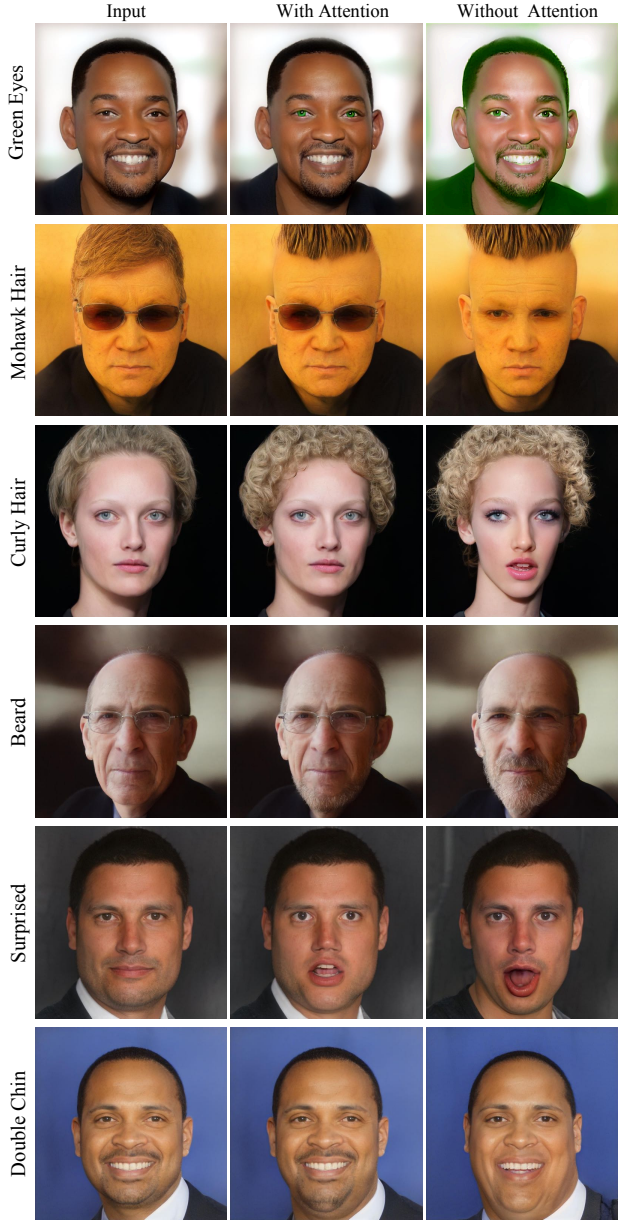}
\end{center}
   \caption{The effectiveness of the attention mechanism. We compare the edited results of our full model (the second column) and the one without the attention module (the third column). The corresponding text prompts are shown on the side.}

\label{fig:attention-analysis}
\end{figure}

\begin{table}[!tb]
  \centering
  \tabcolsep=0.1cm
  \scalebox{0.86}{
  \begin{threeparttable}
  \caption{Quantitative comparison with TediGAN and StyleCLIP by using different metrics. $\downarrow$ indicates that lower is better. $\uparrow$ indicates that higher is better.}
  \label{tab:comparison1}
    \begin{tabular}{cccccccccc}
    \toprule
    \multirow{2}{*}{Attribute}&
    \multicolumn{3}{c}{FID$\downarrow$}&\multicolumn{3}{c}{CS$\uparrow$}&\multicolumn{3}{c}{ED$\downarrow$}\cr
    \cmidrule(lr){2-4} \cmidrule(lr){5-7} \cmidrule(lr){8-10}
    &TG&SC&Ours&TG&SC&Ours&TG&SC&Ours\cr
    \midrule
        Curly hair  & 32.37 & 15.61 & 9.87          & 0.52 & 0.75 & 0.78          & 0.93 & 0.67 & 0.62          \\
        Mohawk hair & 22.78 & 14.81 & 10.02         & 0.49 & 0.75 & 0.81          & 0.96 & 0.68 & 0.59          \\
        Purple hair & 27.17 & 11.13 & 13.43         & 0.65 & 0.82 & 0.78          & 0.81 & 0.40 & 0.62          \\
        Surprised   & 11.73 & 9.31  & 4.73          & 0.68 & 0.89 & 0.96          & 0.78 & 0.43 & 0.27          \\
        Angry       & 14.46 & 6.15  & 2.51          & 0.59 & 0.87 & 0.88          & 0.88 & 0.49 & 0.47          \\
        \midrule
        Average     & 21.70 & 11.40 & \textbf{8.11} & 0.59 & 0.82 & \textbf{0.84} & 0.87 & 0.53 & \textbf{0.51} \\
    \bottomrule

\multicolumn{10}{c}{\fontsize{8}{8} {\textit{Notations: TG - TediGAN; SC - StyleCLIP; FID - Fr\' echet Inception Distance;}}} \cr
\multicolumn{10}{c}{\fontsize{8}{8} {\textit{CS - Cosine Similarity; ED - Euclidean Distance.}}}
    \end{tabular}
    \end{threeparttable}}
\end{table}

\subsection{Learning to Attend}
In this section, we inspect the attention module learned by our framework. Fig.~\ref{fig:attention-map} shows the produced attention maps when editing different facial attributes. We can observe that our approach succeeds in attending to the most relevant facial regions and creating semantic manipulation based on the text descriptions. We can see that in the example of eye shadow editing (the $1^{st}$ column), the attention map has large values in the vicinity of the eyes, as expected. It can also be observed that the learned attention maps can correctly localize the eyes, lip, beard and hair, thus avoiding unnecessary changes to irrelevant areas when performing editing.


\subsection{Qualitative and Quantitative Comparisons}
First, we compare our approach with two recently proposed text-based manipulation methods: TediGAN and StyleCLIP. TediGAN encodes both the image and the text into the StyleGAN latent space and can support text-based image manipulation with CLIP model. StyleCLIP investigates three techniques that combine CLIP with StyleGAN. We use the latent mapper approach, which is closer to our architecture, for comparison.

In addition, we provide comparative results with other prominent StyleGAN manipulation methods: InterFaceGAN and StyleFlow. InterfaceGAN performs linear manipulation in the GAN latent space, while StyleFlow extracts non-linear editing paths in the latent space using conditional continuous normalizing flows.





\paragraph{Qualitative Results.}
Fig.~\ref{fig:compare1} presents the visual comparison with TediGAN and StyleCLIP by using different text descriptions. As can be observed, our method achieves more precise control over facial attributes, including the Mohawk hairstyle and the surprised expression. It can be seen that TediGAN fails to change the hairstyle and could alter the skin color. In addition, the perceived gender (the second example in the rightmost column) changes when performing expression manipulation. StyleCLIP can produce better results than TediGAN and successfully achieve these editing. However, other attributes can be heavily altered. We can observe that there exist noticeable changes of the background when editing the hairstyle and expression. By contrast, with the guidance of the attention map, our model effectively avoids unwanted changes by attending only to the target regions.

In Fig.~\ref{fig:compare2}, we provide a visual comparison with InterFaceGAN and StyleFlow. We consider adding a beard on the faces, which the compared methods succeed in performing. Note that these two approaches require labeled data for supervision, and unlike our method, they cannot apply zero-shot manipulations. We notice that InterFaceGAN and StyleFlow may cause unwanted changes to the eyebrows when asked to add a beard. Compared to these approaches, our method is able to perform more controllable edits by only focusing on the most relevant parts in the images.

\paragraph{Quantitative Results.}
The superiority of our method can also be validated by quantitative evaluation. We evaluate FID, CS and ED based on 10,000 samples randomly generated with the StyleGAN2 generator. The results of five attributes editing compared with TediGAN and StyleCLIP are tabulated in Table~\ref{tab:comparison1}. It can be seen that our method outperforms other approaches in terms of both visual quality (lower FID values) and identity preservation (higher CS and lower ED values). Our results are more favorable because our model can correctly localize the area of interest, thus avoiding unnecessary changes to irrelevant parts.

\begin{figure}[!tb]
\begin{center}
   \includegraphics[width=1\linewidth]{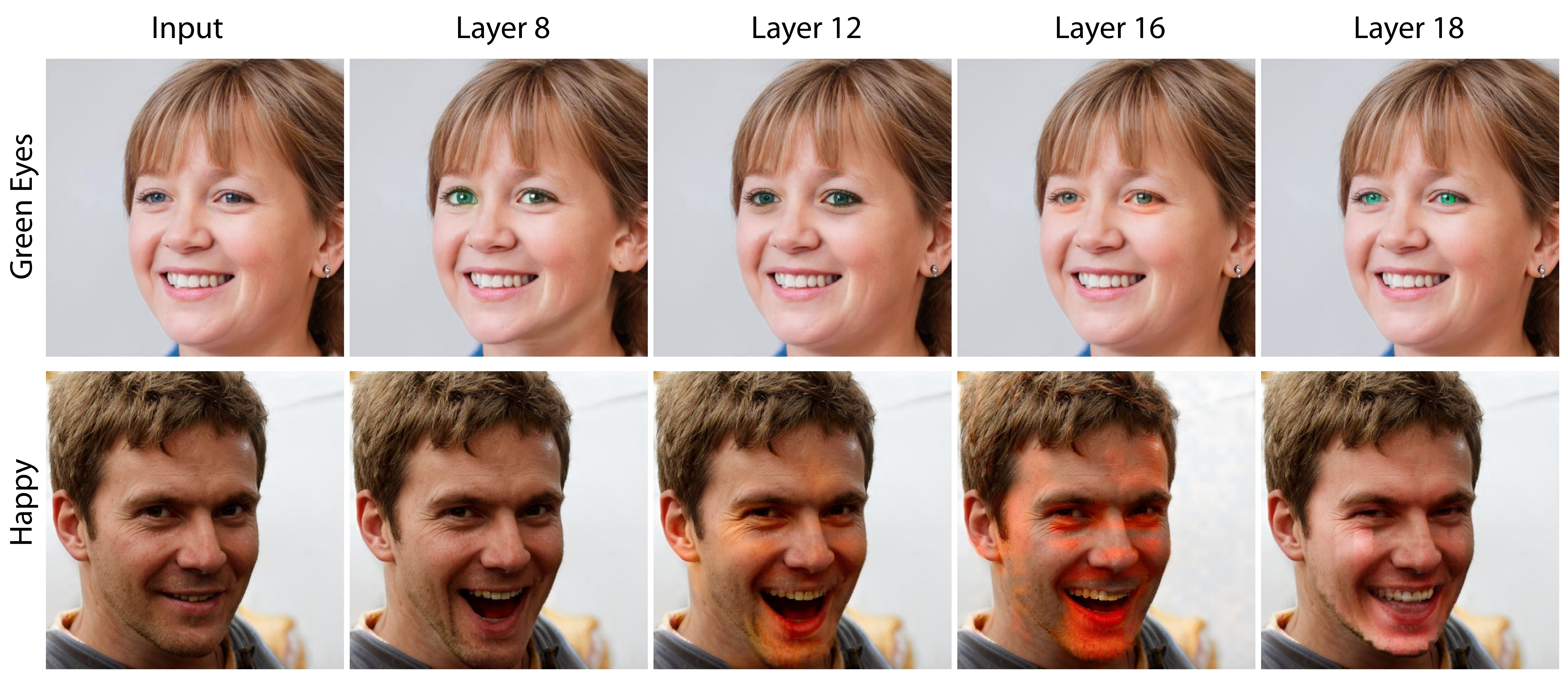}
\end{center}
   \caption{Visual comparison by performing blending in different feature layers. The corresponding text prompts are shown on the side. We can clearly tell that for color manipulation Layer 18 is the best, while for structural manipulation Layer 8 is the best.}
   \label{fig:compare-feature-layers}
\end{figure}

\subsection{Attention Mechanism Analysis}

\paragraph{Attention Module.}

Our attention module is used to produce a probability map that decides the regions to be modified. To demonstrate the effectiveness of the proposed attention mechanism, we perform an ablation study by muting the attention component. We show several examples in Fig.~\ref{fig:attention-analysis}. As can be seen, without the guidance of the attention map, different attributes are entangled. In the example of facial structure manipulation, the eyeglasses and face shape (the $2^{nd}$ and $3^{rd}$ row) are entangled with the hairstyle. We can also see obvious changes of the background and clothes when editing the surprised expression (the $5^{th}$ row). Moreover, the entanglement is particularly noticeable when performing color manipulation, and we can see that the background color and skin tone change significantly when we try to achieve green eyes (the $1^{st}$ row). By contrast, with the guidance of the attention map, our algorithm can effectively avoid unwanted changes by attending only to the target regions.


\paragraph{Blending in Different Layers.}
In this part, we explore the effect of feature blending in different layers. Two examples are shown in Fig.~\ref{fig:compare-feature-layers}. As can be observed, our approach performs well on the color prompt (the $1^{st}$ row) when performing the blending in the highest layer (Layer 18), and blending in the lower layer (Layer 8) results in changes of the eye shape. By contrast, for facial structural editing like the happy expression (the $2^{nd}$ row), it is better to adopt lower layers for feature blending. This is because the fact that StyleGAN controls high-level aspects such as the pose, hair style and face shape in the coarse and middle layers ($4^2 - 32^2$) and deals with colors and smaller details in the finer layers ($64^2 - 1024^2$).

\begin{figure}[!tb]
\begin{center}
   \includegraphics[width=0.98\linewidth]{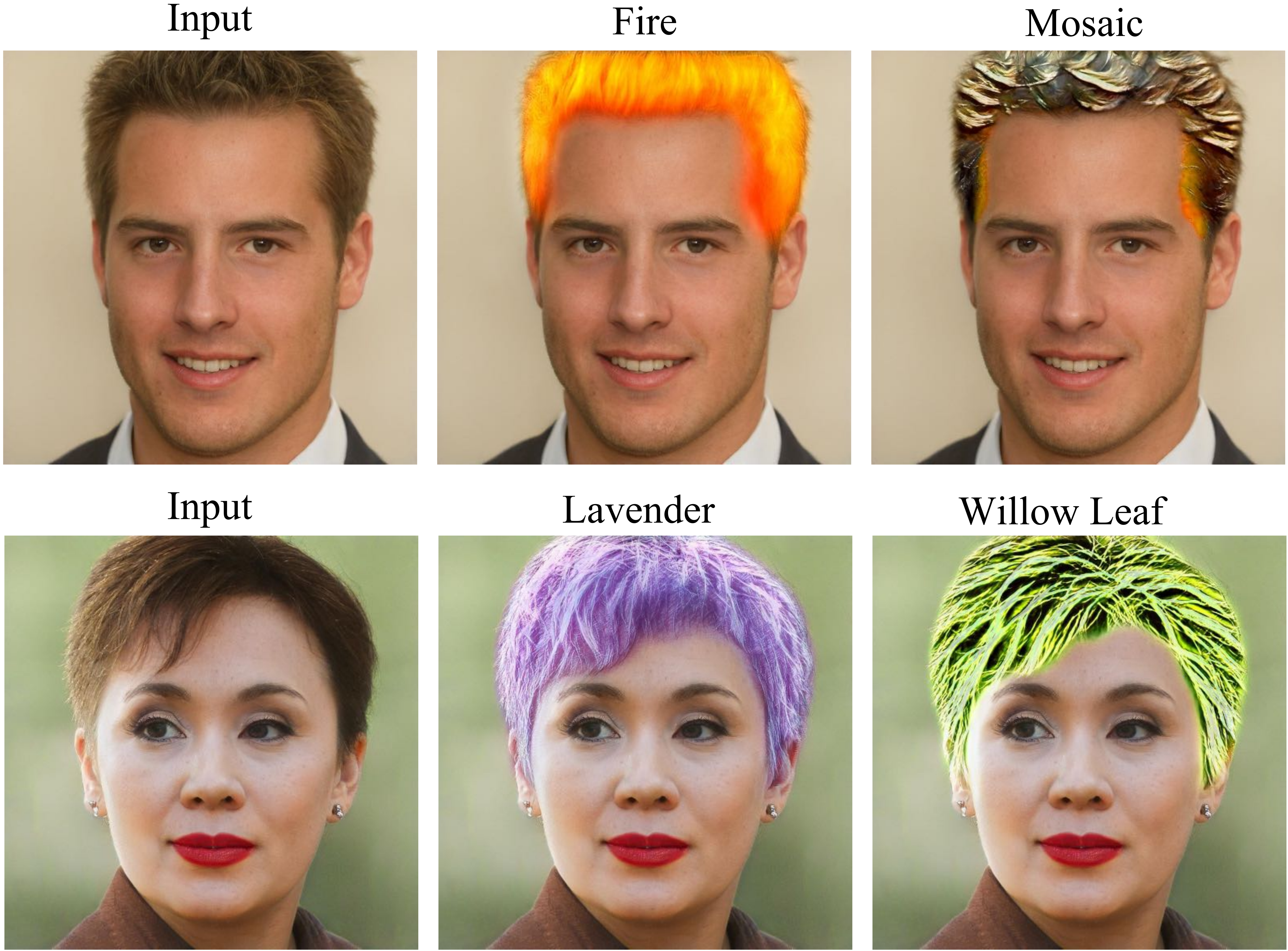}
\end{center}
   \caption{Two-step image manipulation. In the first step, we train the attention module with the text prompt ``Black Hair''. In the second step, the attention map is fixed, and the mapper is trained with the text prompts ``Fire'', ``Mosaic'', ``Lavender'' and ``Willow Leaf'',  respectively.}
\label{fig:two-step}
\end{figure}

\paragraph{Two-Step Image Manipulation.}
In addition to simultaneously learning the attention masks and performing various face editing, we further explore a two-step training approach, where we use the first prompt to define the mask and the second one for manipulation with the mask produced in the first step. In this way, we are able to control facial regions with more interesting descriptions. Fig~\ref{fig:two-step} shows four examples. In particular, we first use the  ``Black Hair'' prompt to extract the hair area and fix the obtained mask to train models using the text prompt ``Fire'', ``Mosaic'', ``Lavender'' and ``Willow Leaf'', respectively. In this experiment, we perform the feature blending at the $12^{th}$ layer. As can be seen, by explicitly limiting the area that can be edited, our model can achieve more meaningful manipulation in the specific facial region. These results demonstrate that the attention mechanism can assist with an intuitive control over facial areas using text descriptions.

\section{Conclusion and Future Work}
\label{sec:future}

We presented a novel method for semantic editing of facial regions guided by text descriptions. Our approach features a novel attention framework that leverages the knowledge learned from text-image joint embedding to guide the image generation process of StyleGAN. The attention module learns attention maps with text supervision only and is particularly useful to focus the manipulation on the intended regions of interest.
Unlike previous methods that strongly rely on the disentanglement of different attributes in the latent space, our approach alleviates this dependency by explicitly restricting the altered spatial regions.
Experiments demonstrate the superior performance of our method over previous works.

In the future, it will be interesting to explore the potential of these attention masks as means to segment facial regions. Using guiding text or possibly a few-shot, our mapper can detect the corresponding facial regions. Such segmentation may successfully be generalized to various faces with different poses. Furthermore, we believe that such attention mechanisms can be extended beyond the facial domain and bypass attributes that are spatially entangled.

{\small
\bibliographystyle{ieee_fullname}
\bibliography{egbib}
}

\end{document}


\title{FEAT: Face Editing with Attention \\ (Supplementary Material)}
\author{
  Xianxu Hou\textsuperscript{1},
  Linlin Shen\textsuperscript{1},
  Or Patashnik\textsuperscript{2},
  Daniel Cohen-Or\textsuperscript{2},
  Hui Huang\textsuperscript{1} \\
  \textsuperscript{1}Shenzhen University \quad
  \textsuperscript{2}Tel Aviv University \\
}
\maketitle

    \section{Qualitative Comparisons}
    Here, we provide additional comparisons of our method with the current state-of-the-art methods for text-based manipulation: TediGAN~\cite{xia2021tedigan} and StyleCLIP~\cite{patashnik2021styleclip}. These methods also combine CLIP with StyleGAN~\cite{karras2019style,karras2020analyzing} to achieve text-based editing. For StyleCLIP, we use the latent mapper approach for comparison, which is closer to our architecture. For structural manipulation, we also compare our method with StyleFusion~\cite{kafri2021stylefusion}, which provides fine-grained control over each region of the generated image by fusing different latent codes. In particular, we first use StyleCLIP to manipulate the latent code to obtain the edited representation, which is then fed to StyleFusion hierarchy to control the areas relevant to the desired edit. Then, we use the original latent code to control the remaining facial areas. 
    
\begin{figure*}[!bt]
    \begin{center}
       \includegraphics[width=0.9\linewidth]{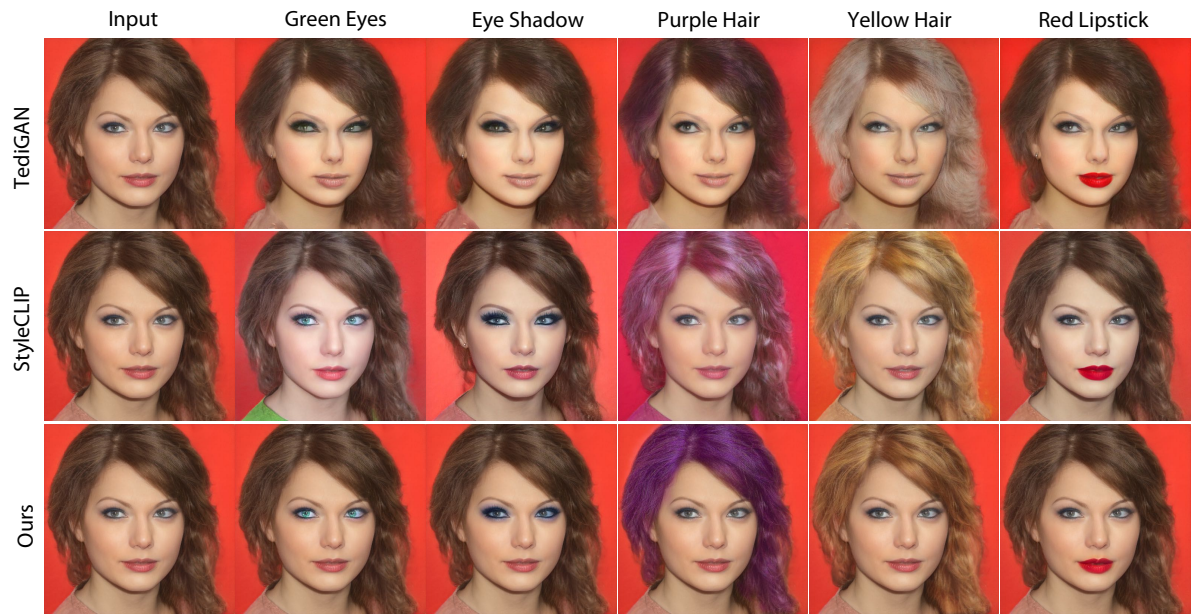}
    \end{center}
       \caption{Visual comparison of our method with TediGAN and StyleCLIP. The driving descriptions are indicated above each column.}
    \label{fig:supp0}
    \end{figure*}

    Fig.~\ref{fig:supp0} presents the visual comparison using different color prompts. As can be seen, our method can achieve more precise control over various facial attributes, including different colors of eyes, hair, lipstick and eye shadow. We observe that TediGAN fails to produce colorful eyes and hair, and the perceived identity is changed when editing lipstick. StyleCLIP can produce better results than TediGAN and successfully change the colors of different facial attributes. However, the color of the background and clothes may also be altered when producing the green eyes.

\begin{figure*}[!bt]
    \begin{center}
       \includegraphics[width=0.8\linewidth]{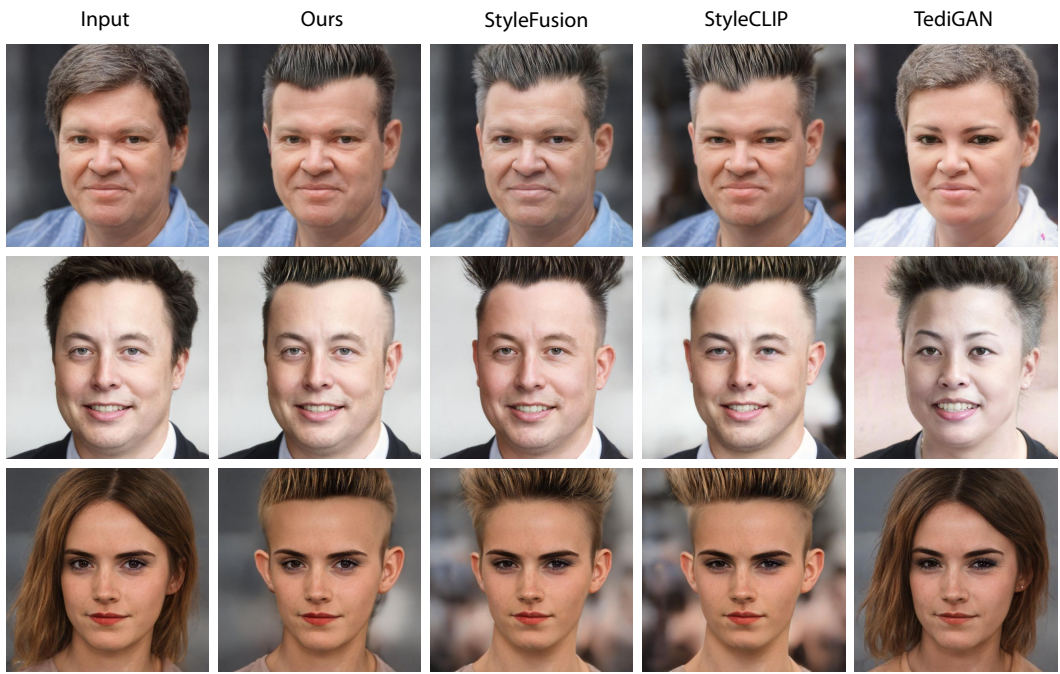}
    \end{center}
       \caption{Visual comparison of our method with TediGAN, StyleCLIP and StyleFusion on Mohawk hairstyle. 
       As can be seen, our method yields a disentangled edit. For example, in the last row the background is unnecessarily modified even when combining StyleCLIP with StyleFusion.}
    \label{fig:supp-mohawk}
    \end{figure*}

\begin{figure*}[!bt]
    \begin{center}
       \includegraphics[width=0.8\linewidth]{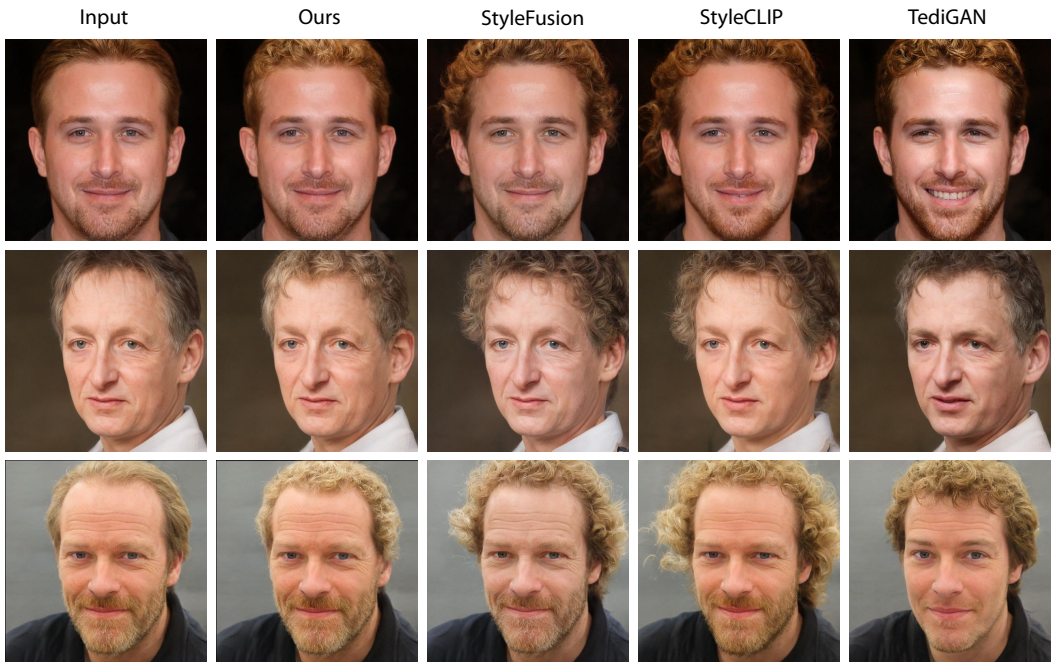}
    \end{center}
       \caption{Visual comparison of our method with TediGAN, StyleCLIP and StyleFusion on curly hairstyle. 
       As can be seen, while in other methods the hair style (curly) is entangled with the hair length, our method successfully alters only the hair style.} 
    \label{fig:supp-curly}
    \end{figure*}

\begin{figure*}[]
    \begin{center}
       \includegraphics[width=0.8\linewidth]{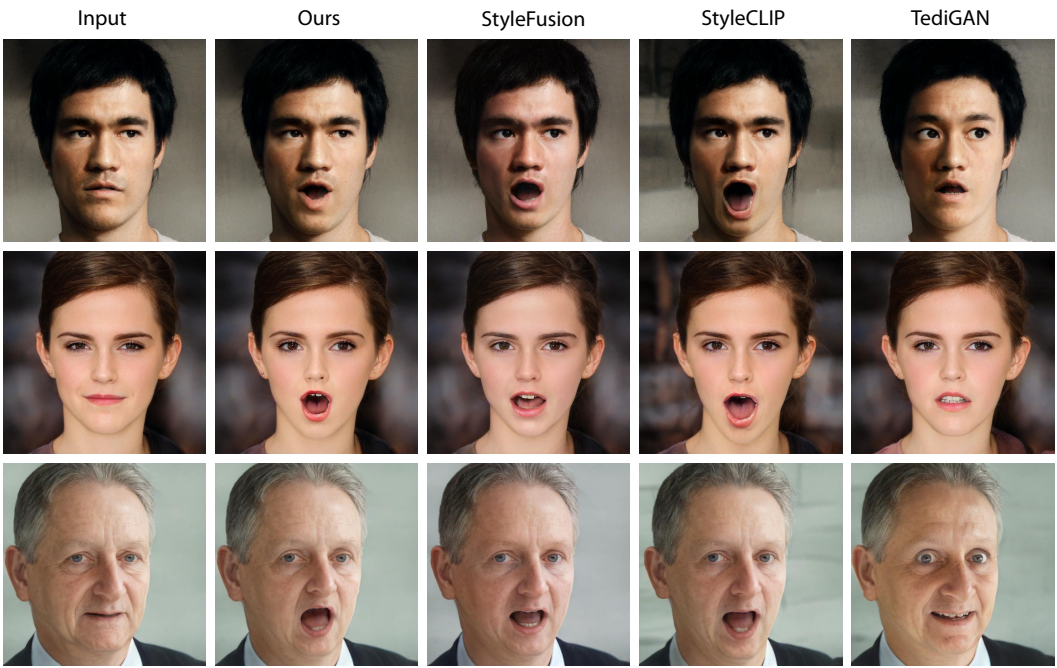}
    \end{center}
       \caption{Visual comparison of our method with TediGAN, StyleCLIP and StyleFusion on surprised expression. 
       As can be seen, our method achieves more faithful results compared to other works. For example, our method is more disentangled than StyleCLIP, which slightly changes the identity, and more accurate than StyleFusion, which does not alter the eyes.}
    \label{fig:supp-surprised}
    \end{figure*}

\begin{figure*}[]
    \begin{center}
       \includegraphics[width=0.8\linewidth]{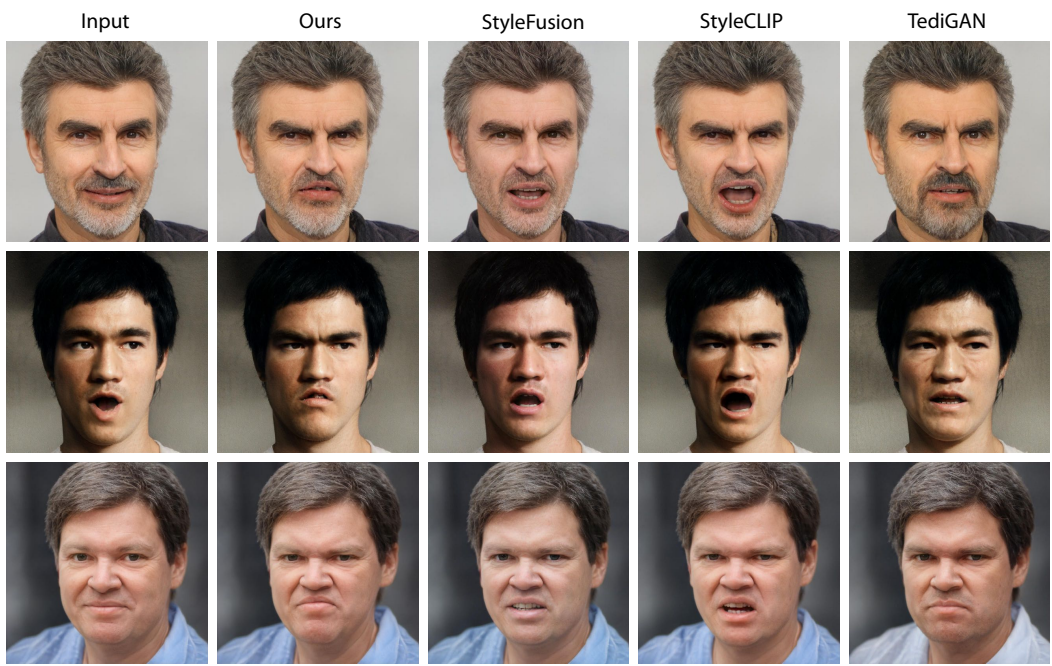}
    \end{center}
       \caption{Visual comparison of our method with TediGAN, StyleCLIP and StyleFusion on angry expression. For StyleFusion, we only edit the relevant attributes: face, eyes and mouth.}
    \label{fig:supp-angry}
    \end{figure*}

  Figs.~\ref{fig:supp-mohawk},~\ref{fig:supp-curly},~\ref{fig:supp-surprised} and~\ref{fig:supp-angry} present the comparisons on several structural manipulations, including different hairstyles and facial expressions. It can be seen that TediGAN struggles in producing the Mohawk hairstyle and the perceived gender has been changed (Fig.~\ref{fig:supp-mohawk}). Although StyleCLIP can successfully achieve these edits, background can be entangled with hairstyle (Fig.~\ref{fig:supp-mohawk}) and expression (Fig.~\ref{fig:supp-surprised}). StyleFusion can better preserve the background than StyleCLIP, however we can observe the tone change of the entire image (Fig.~\ref{fig:supp-surprised}) and slight entanglement between the hair and eyes. By contrast, our approach is able to effectively avoid unwanted changes by providing precise local control over different attributes. 
  Moreover, in StyleFusion the regions are pre-defined independently to the text-prompt, while in our method the regions are learned for the specific text-prompt.

{\small
\bibliographystyle{ieee_fullname}
\bibliography{egbib}
}